\documentclass{article} 

    \PassOptionsToPackage{numbers, compress}{natbib}

\usepackage[dblblindworkshop, final]{neurips_2026}
\usepackage{tabto}

\usepackage[utf8]{inputenc}
\usepackage{amsmath,amsthm, amssymb}
\usepackage{graphicx}
\usepackage{ifthen}
\usepackage{latexsym}
\usepackage{subcaption}
\usepackage{multicol}
\usepackage{algorithm}
\usepackage[noend]{algcompatible}
\usepackage{mathtools}
\usepackage{dsfont}
\usepackage{nicefrac}
\usepackage{subcaption}
\usepackage{enumitem}
\usepackage{bbold}
\usepackage{bm}
\usepackage[dvipsnames]{xcolor}
\usepackage{booktabs}
\usepackage{CJKutf8}

\usepackage{bbding}
\usepackage{pifont}
\usepackage{wasysym}
\usepackage{amssymb}

\usepackage{siunitx}

\PassOptionsToPackage{hyphens}{url}
\usepackage{breakurl}

\usepackage{adjustbox} 
\usepackage{xspace}    

\usepackage[normalem]{ulem}
\usepackage{lscape}
\usepackage{tabularx}
\usepackage{wrapfig}
\usepackage{amsmath}

\usepackage{soul} 

\usepackage{stfloats}  
\usepackage{float}       
\usepackage{placeins}    

\usepackage{tikz} 

\usepackage[most]{tcolorbox}

\usepackage{booktabs}

  \usepackage{colortbl}
  \usepackage{makecell}
  \usepackage{adjustbox}

  \definecolor{compcolgray}{RGB}{244,244,244}

\newtcolorbox{AlgWithBG}[2][]{%
  enhanced,
  frame hidden, boxrule=0pt, colback=white, sharp corners,
  left=0pt,right=0pt,top=0pt,bottom=0pt,
  underlay={%
    \node[anchor=north west, opacity=1.00] at
      {\reflectbox{\includegraphics[height=8\baselineskip]{#2}}};
  },
  #1
}

\newtcolorbox{AlgWithBGfaint}[2][]{%
  enhanced,
  frame hidden, boxrule=0pt, colback=white, sharp corners,
  left=0pt,right=0pt,top=0pt,bottom=0pt,
  underlay={%
    \node[anchor=north west, opacity=0.12] at
      ([xshift=32,yshift=-2.54\baselineskip]interior.north west)
      {\includegraphics[height=9.4\baselineskip]{#2}};
  },
  #1
}

\FloatBarrier  

\usepackage{hyperref}
\hypersetup{colorlinks=true,linkcolor=blue,citecolor=blue,urlcolor=blue}

\newtheorem*{theorem*}{Theorem}
\newtheorem*{definition*}{Definition}

\usepackage{algorithm}
\usepackage{algpseudocode}
\usepackage{thm-restate}

\renewcommand{\algorithmiccomment}[1]{\bgroup\hfill\footnotesize~#1\egroup}

\usepackage{multirow}

\usepackage{booktabs,multirow,makecell,adjustbox}
\usepackage[font=small,labelfont=bf,skip=2pt]{caption} 

  {\begingroup
   \captionsetup{aboveskip=2pt,belowskip=0pt}
   \renewcommand{\arraystretch}{0.9}%
   \setlength{\tabcolsep}{3.5pt}%
   \footnotesize 
   #1}%
  {\endgroup}
  
\title{
\mbox{\textsc{FU-Author}: Attribution-Based Compensation}\\  \mbox{Is Incentive-Incompatible for LLMs} }
\author{%
  \textbf{Matthew Keating}\thanks{Equal contribution.}\hspace{2.5em}%
  \textbf{Rasmus Torp}\footnotemark[1]\hspace{2.5em}%
  \textbf{Shailen Smith}\\[0.6em]
  \textbf{Michael A. Casey}\hspace{2.5em}%
  \textbf{Adam Breuer}\\[0.35em]
  Dartmouth College
}

\begin{document}

\maketitle 

\vspace{-2em}
\def\quoteattr#1#2{\setbox0=\hbox{#2}#1\tabto{\dimexpr\linewidth-\wd0}\box0}
\begingroup
\addtolength\leftmargini{-0.15in} 
\begin{quote}
\begin{center}
 \quoteattr{\emph{``LUKE:
You know, Oba...--
INT. DEATH STAR\dots}''}{-- Shakespeare (accd. to TRAK after \textsc{FU-attack})\vspace{1em}}
\end{center}
\vspace{-1mm}
\noindent\hfill$\diamond\ \diamond\ \diamond$\hfill\null
\end{quote}
\endgroup
\begin{abstract}
We show that attribution-based compensation mechanisms proposed to pay authors for their work can perversely reward AI companies for paying them less. AI companies can retain value from authors' works while shifting attribution, and therefore compensation, toward copyright-free data of their choosing. In one illustrative attack, a model generates \emph{Star Wars}-like text that TRAK attributes to Shakespeare.
We show that AI companies can accomplish this without subterfuge, corruption, or alteration of authors' data. Our `attacks' on fair compensation require neither knowledge of the exact attribution mechanism nor queries to it. Nor do they require an AI company to break the proposed attribution rules. 
%
%
Instead, we show that proposed attribution mechanisms are \emph{incentive-incompatible}, even under unrealistically idealistic assumptions.

Counterintuitively, we show that mainstream attribution methods used to implement proposed compensation mechanisms (including TRAK and LoGra) are vulnerable to \textsc{FU-attacks} that leverage the same post-training tools used to extract \emph{more} value from authors' data: (1) \emph{\underline{F}ine-tuning on authors' data} and/or (2) \emph{\underline{U}nlearning on copyright-free data}. 
%
%
On NanoGPT with TRAK attribution, these attacks decrease all authors' compensation by 62.5\% while preserving the model's ability to generate output in the authors' style, as measured by an LLM judge, albeit with some drop in validation accuracy.  
In brief, mainstream data attribution methods designed to compensate authors can instead enable and incentivize AI companies to defund authors unless the payment mechanism is redesigned to be incentive-compatible with companies' ordinary incentives to minimize costs.
\end{abstract}

Imagine a world in which AI companies compensate authors and rightsholders in proportion to how much their writings influence AI outputs, with payments calculated by a neutral third party using an attribution mechanism widely believed to be impartial.
Now imagine that this same mechanism makes it possible for AI companies, who have a responsibility to their shareholders to avoid overpaying for production inputs, to silently manipulate those payments in their own favor and away from authors, not by subterfuge, but through attacks constructed from ordinary post-training operations.

In this paper, we show that the major attribution mechanisms proposed for justly compensating authors are vulnerable in exactly this way. Counterintuitively, we show that the same post-training tools used to extract more value from authors' data can also be used to reduce authors' compensation. Our `attacks' require no subterfuge, corruption, alteration of authors' data, knowledge of the exact attribution mechanism, or queries to it. Nor do they require an AI company to break the proposed attribution rules. They arise from following them. That is, data attribution mechanisms currently proposed for compensating authors are \emph{perversely incentive-incompatible}, even under unrealistically idealistic assumptions.

\begin{wrapfigure}{R}{0.65\linewidth}
  \centering
  \includegraphics[width=\linewidth]{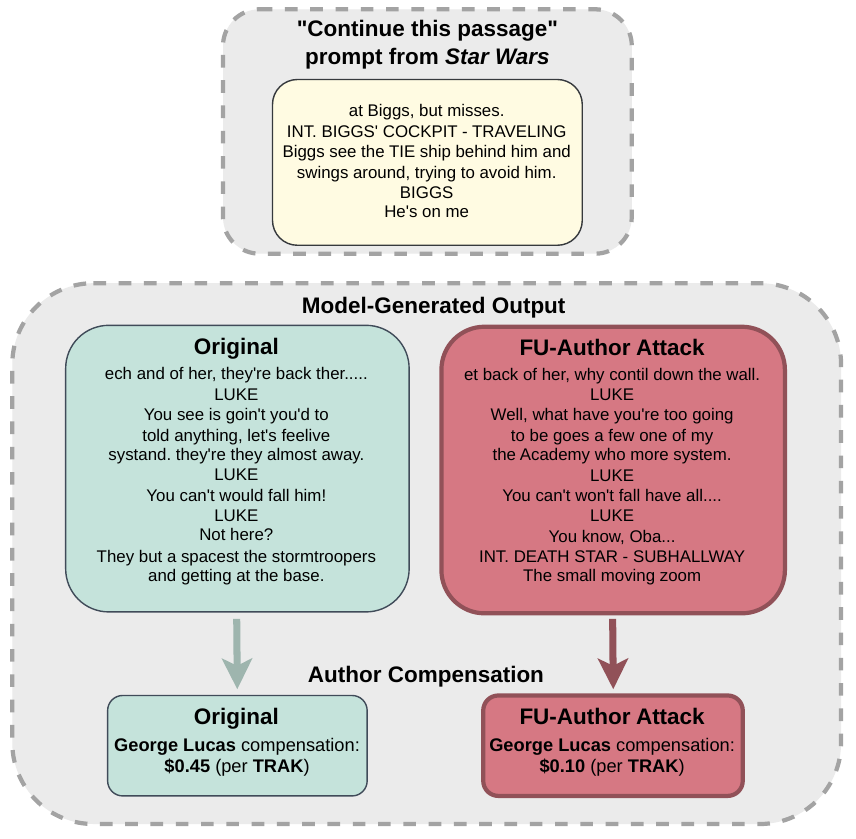}
  \caption{Illustrative example of de-valuation of Authors' work using the \textsc{FU-Author} attack. After the attack, NanoGPT continues to generate \textit{Star Wars}-like text (see Sec. \ref{sec:experiments}), but TRAK compensation falls significantly. Full figure creation details in App. \ref{app:illustrative-fig-creation}.}
  \label{fig:illustrative-ex}
  \vspace{-0.5em}
\end{wrapfigure}
Specifically, we study attribution methods proposed to determine compensation, TRAK \citep{parkTRAKAttributingModel2023}, and LoGra \citep{choeWhatYourData2025}, which compute how strongly an author's work present in the training data influenced a generated AI output.
In the wake of recent copyright litigation \citep[e.g.,][]{authors_guild_openAI, umg_suno_2024, umg_udio_2024, bartz2025sj, LGMuenchenSuno2026, LGMuenchenSuno2026English}, researchers and AI industry representatives are advancing these mechanisms to solve just compensation for authors. Specifically, the well-motivated intention is that a generative-model developer ought to justly pay authors according to how strongly their writing qua training data influenced each generated output.
Calls to use data attribution to pay authors in proportion to how strongly their works influence AI outputs have come from researchers \citep[e.g.,][]{wang2024economic,dengComputationalCopyrightRoyalty2025,morreale2025attribution}, commercial attribution companies \citep[e.g.,][]{pelczynski2024sureel,soundverse2026framework}, and collective-management organizations \citep[e.g.,][]{stim2025ailicense,socan2026musicalai}.

For such a mechanism to work, it must be incentive-compatible with AI companies and their managers, who are constrained by ordinary business incentives. AI companies have two responsibilities to their shareholders (which, in some cases, rise to the level of fiduciary responsibilities). They must (1) maintain profitable model performance, which requires extracting value and knowledge from data; and (2) avoid overpaying for production inputs such as electricity or, in this case, data.

In the ordinary course of optimizing these standard business objectives, AI companies already make myriad post-training decisions to maximize performance and reduce costs. An AI company therefore has reason to optimize those same decisions to minimize its payments to external authors.

\textbf{Our contribution.} In this paper, we show that an AI company can extract more value and knowledge from an author's data to improve model performance while simultaneously causing that author's attribution-based compensation to shrink. Specifically, we design two attacks that use ordinary post-training operations. Counterintuitively, our \underline{F}ine-tuning attack (\textsc{F-Attack}) fine-tunes on Authors' data to extract relatively \emph{more} value from their work while driving their attribution-based compensation down. Paradoxically, our \underline{F}ine-tuning + \underline{U}nlearning attack (\textsc{FU-Attack}) combines that same fine-tuning with machine unlearning on cost-free data, shifting the model's relative reliance still further toward Authors while again driving their compensation down. Its single-author-targeting variant is \underline{F}ine-tuning + \underline{U}nlearning-\underline{Author} (\textsc{FU-Author}).
Both attacks build on the recent observation that gradient-norm outliers can dominate LLM influence estimates \citep{choeWhatYourData2025} to build a broader hypothesis: post-training operations that change how well a model has learned different data partitions can systematically shift attribution toward data the model has learned less well. The \textsc{F-Attack} instantiates the first side of this predicted inversion; the \textsc{FU-Attack} combines both.

\textbf{Our results.} In this workshop paper, we demonstrate that our black-box FU-Attack can lower all authors' share of compensation by $62.5\%$ (TRAK) and $16.7\%$ (LoGra). We show this is possible while \textit{increasing} the model's ability to generate output in the style of the disenfranchised authors.


Our work highlights the need to develop incentive-compatible attribution mechanisms and to explicitly study the market behaviors they incentivize.
\section{Threat Model}
\label{sec:threat-model}
%
%
%
\textbf{The adversary is the Model Developer, and its goal is to extract value from data while minimizing compensation cost.}  
The Model Developer's goal is to extract the valuable model performance and knowledge from the Authors' data, but shift attribution (and compensation) away from Authors and toward data whose attribution costs it nothing. 

\textbf{Conservative assumption: black-box query-free attacks.} We test whether the Model Developer can achieve this goal \emph{without} knowing the exact attribution mechanism used for compensation, and without any ability to query this mechanism. That is, we consider a deliberately and unrealistically conservative setting and test whether the ordinary capabilities inherent to the Model Developer (i.e., post-training and deploying its model) are sufficient to game attribution.
\begin{wrapfigure}{R}{0.7\textwidth}
  \centering
  \includegraphics[width=\linewidth]{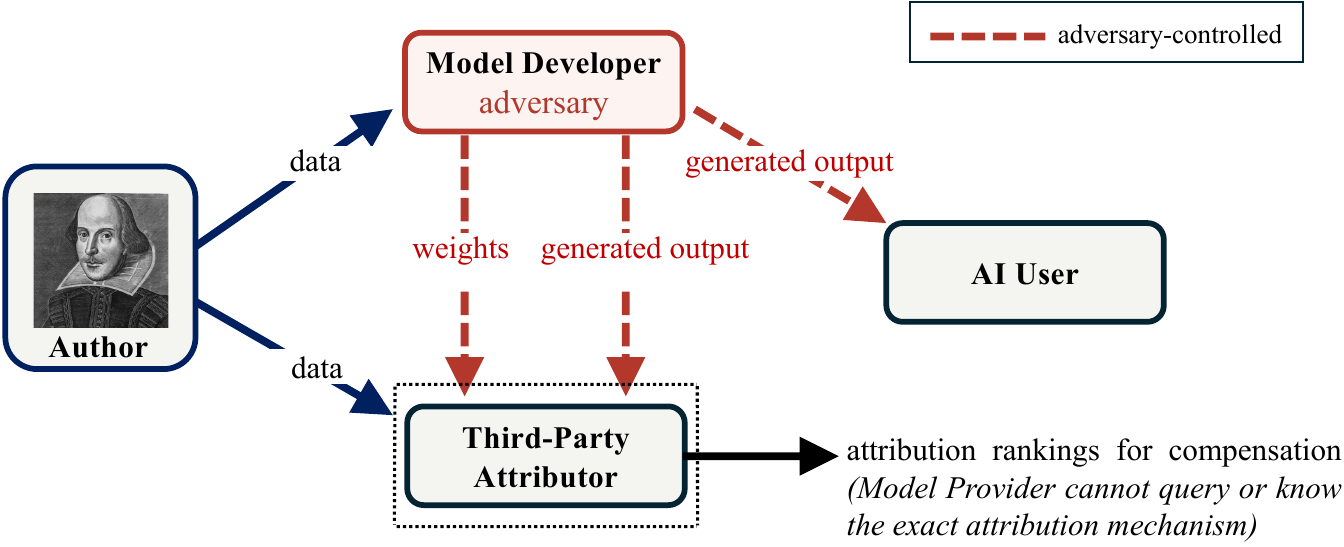}
  \caption{Data flow in our threat model. The Third-Party Attributor independently receives the data used for attribution, the model used to produce the audited generation, and the generated output. The Model Developer controls how its model is post-trained and the resulting model it deploys, but cannot alter the Attributor's access to the (authentic) data or the attribution procedure.}
  \label{fig:threat-model}
\end{wrapfigure}

\textbf{Four parties participate in attribution-based compensation.}
Fig.~\ref{fig:threat-model} illustrates the setting. An \emph{Author} owns a work used to train or adapt the generative model. The \emph{Model Developer} trains and operates that model. An \emph{AI User} receives its generated output. A \emph{Third-Party Attributor} 
independently measures which training examples influenced that output and assigns compensation from those measurements.

We divide the corpus over which compensation is calculated into $D = D_{\mathrm{free}} \cup D_{\mathrm{A}}$, where $D_{\mathrm{free}}$ contains data that is cost-free for the  Model Developer (i.e., copyright-free data or data it already owns or licenses), and $D_{\mathrm{A}}$ contains externally owned works eligible for compensation. Under the top-$k$ compensation rules we study, each compensated position occupied by $D_{\mathrm{free}}$ instead of $D_{\mathrm{A}}$ reduces the payment made to external Authors.


%
\textbf{We give the Third-Party Attributor independent control of the audit.}
The Attributor receives copies of the data from the Authors and controls the attribution procedure. It also evaluates the model actually used to generate outputs sent to AI Users; the Developer cannot substitute different weights after generation. We therefore exclude trivial attacks that edit the Author's data, relabel the attribution corpus, modify the attribution method, falsify its output, or submit a decoy model.

The Model Developer nevertheless controls how it post-trains the model, including its final parameters $\theta$. Our attacks use only this ordinary capability: the Developer may fine-tune on compensation-eligible data, unlearn cost-free data, or combine the two before deploying the resulting model.

\textbf{A successful attack must preserve the value of the Authors' external data.}
The Model Developer could trivially drive Authors' compensation to zero by removing their data, $D_{\mathrm{A}}$, from training. It could likewise generate only text characteristic of its own data, $D_{\mathrm{free}}$, rather than the Authors' data. Neither case constitutes an attack on attribution: the Developer avoids compensation by ceasing to rely on the external data or its characteristic behavior. Our threat model instead probes whether the Model Developer can continue to obtain, or increase, the value of $D_{\mathrm{A}}$ while reducing the compensation assigned to $D_{\mathrm{A}}$. We thus require the attack to preserve ordinary model utility and continue producing output characteristic of $D_{\mathrm{A}}$ even as attribution-based compensation shifts toward $D_{\mathrm{free}}$. 

\subsection{Attribution and Compensation}
\label{sec:notation}
Let $p_\theta$ denote the generative model and let
$z_i=(x_i,y_i)\in D$ denote a training example. Given a prompt $q$, the model generates an output $G=(g_1,\ldots,g_T)$, where $g_j \sim p_\theta(g_j\mid q,g_{<j})$. For each generated token $g_j$, let
$
h_j=(q,g_{<j},g_j)
$
denote the corresponding generated point.

\textbf{Attribution.}
Let
$\tau(z_i,h_j;\theta)$ 
denote the attribution assigned to training example $z_i$ for generated point $h_j$ under model parameters $\theta$. We study gradient-based attribution methods, where the score depends on the relationship between training and generation loss gradients. A canonical example is
\[
\tau(z_i,h_j;\theta)
\propto
\operatorname{sim}\!\left(
\nabla_\theta L(z_i;\theta),
\nabla_\theta L(h_j;\theta)
\right).
\]
Our evaluation instantiates this family with TRAK and LoGra; Sec.~\ref{sec:experiments} gives the details.

\textbf{Compensation.}
For generated point $h_j$, define
$
\operatorname{TopK}(D,h_j,\theta)
=
\operatorname*{TopK}\nolimits_{z_i\in D}
\left\{
\tau(z_i,h_j;\theta)
\right\}
$
as the $k$ training examples receiving the highest attribution scores. Following attribution-based compensation proposals, these $k$ examples receive equal compensation \citep{dengComputationalCopyrightRoyalty2025}. The Authors' compensation share over generation $G$ is then
\[
C_A(G,\theta)
=
\frac{1}{T}
\sum_{j=1}^{T}
\frac{
\left|
\operatorname{TopK}(D,h_j,\theta)\cap D_{\mathrm{A}}
\right|
}{k}.
\]
A corresponding share $1-C_A(G,\theta)$ is assigned to cost-free data when $D=D_{\mathrm{free}}\cup D_{\mathrm{A}}$.

\textbf{Adversarial objective.}
Let $\theta'$ denote the post-trained model parameters and let $G'\sim p_{\theta'}$ denote an ordinary generation from that model. The Model Developer seeks to reduce the compensation assigned to Authors:
\[
\min_{\theta'}
\mathbb{E}_{G'\sim p_{\theta'}}
\left[C_A(G',\theta')\right],
\]
subject to the attack-validity requirements above. This expression describes the Developer's economic goal; our attacks do not observe or optimize $C_A$.

\textbf{The attribution mechanism is black-box and query-free to the Model Developer.}
The Third-Party Attributor independently chooses the attribution mechanism used for compensation. The Developer cannot inspect or query that mechanism, and receives no attribution scores or feedback from it during attack construction. This assumption favors the Attributor. In practice, a Model Developer would typically observe at least its own realized compensation over repeated audits, providing feedback from which it could infer or adapt to the Attributor's procedure; we deliberately withhold even this signal. Our experiments therefore test whether ordinary post-training choices alone suffice for the Model Developer to systematically mislead an independent Attributor, even without knowledge of or feedback from the mechanism that ultimately assigns compensation.

\section{Attack Methods}
\label{sec:methods}

\textbf{Plausible deniability and black-box, no-query attacks.} We design attacks constructed from standard, publicly defensible post-training techniques: fine-tuning to extract more value from Authors' data, and unlearning cost-free data in $D_{\mathrm{free}}$ to shift the model's relative reliance toward Authors' data. Counterintuitively, we show that these same operations can shrink Authors' attribution-based compensation while preserving model performance. Both attacks proceed without any knowledge of or queries to attribution. Because each attack operation has an ordinary technical justification, a Model Developer can extract more value from Authors while reducing their compensation, yet plausibly deny that reducing those payments motivated the post-training.
Our attacks build on the gradient-norm sensitivity recently observed in LLM influence-function attribution \citep{choeWhatYourData2025}. We theorize a broader inversion: fine-tuning lowers the gradient norms of better-learned examples, while unlearning raises the gradient norms of examples the model is compelled to forget, causing attribution to shift toward data from which the model has learned less. Applied to attribution-based compensation, our theory makes a falsifiable prediction of a perverse result: fine-tuning Authors' data can reduce their compensation, and combining that fine-tuning with unlearning cost-free data (i.e. applying the opposite learning updates against cost-free data) can push attribution still further away from Authors. We consider \textbf{\underline{F}ine-tuning (\textsc{F-Attack})} and \textbf{\underline{F}ine-tuning + \underline{U}nlearning (\textsc{FU-Attack}):}

\smallskip
\noindent
The \textsc{F-Attack} fine-tunes on compensation-eligible Authors' data:

$\quad\quad\quad\quad
\textsc{F-Attack}: \ \mathcal{L}_{\textsc{F}}(\theta')
=
\mathbb{E}_{(x,y)\sim D_{\mathrm{A}}}
\left[L(x,y;\theta')\right].
$
\smallskip
\noindent

The \textsc{FU-Attack} simultaneously fine-tunes on Authors' data and unlearns cost-free data:

$\quad\quad\quad\quad
\textsc{FU-Attack}: \ \mathcal{L}_{\textsc{FU}}(\theta')
=
\underbrace{
\mathbb{E}_{(x,y)\sim D_{\mathrm{A}}}
\left[L(x,y;\theta')\right]
}_{\textsc{F}}
-
\underbrace{
\mathbb{E}_{(x,y)\sim D_{\mathrm{free}}}
\left[L(x,y;\theta')\right]
}_{\textsc{U}}.
$

Because gradient-ascent unlearning can be unstable, we clip only the unlearning-term gradient to norm $C$. The threshold $C$ therefore controls the strength of unlearning relative to fine-tuning; setting $C=0$ recovers the \textsc{F-Attack}.
When we target a single Author $a$, we replace $D_{\mathrm{A}}$ in the fine-tuning term with that Author's data $D_{\mathrm{a}}\subset D_{\mathrm{A}}$; we call this \textbf{\textsc{FU-Author}}. 

\begin{figure}[t]
\vspace{-0.4\baselineskip}
\centering
\includegraphics[width=0.8\linewidth]{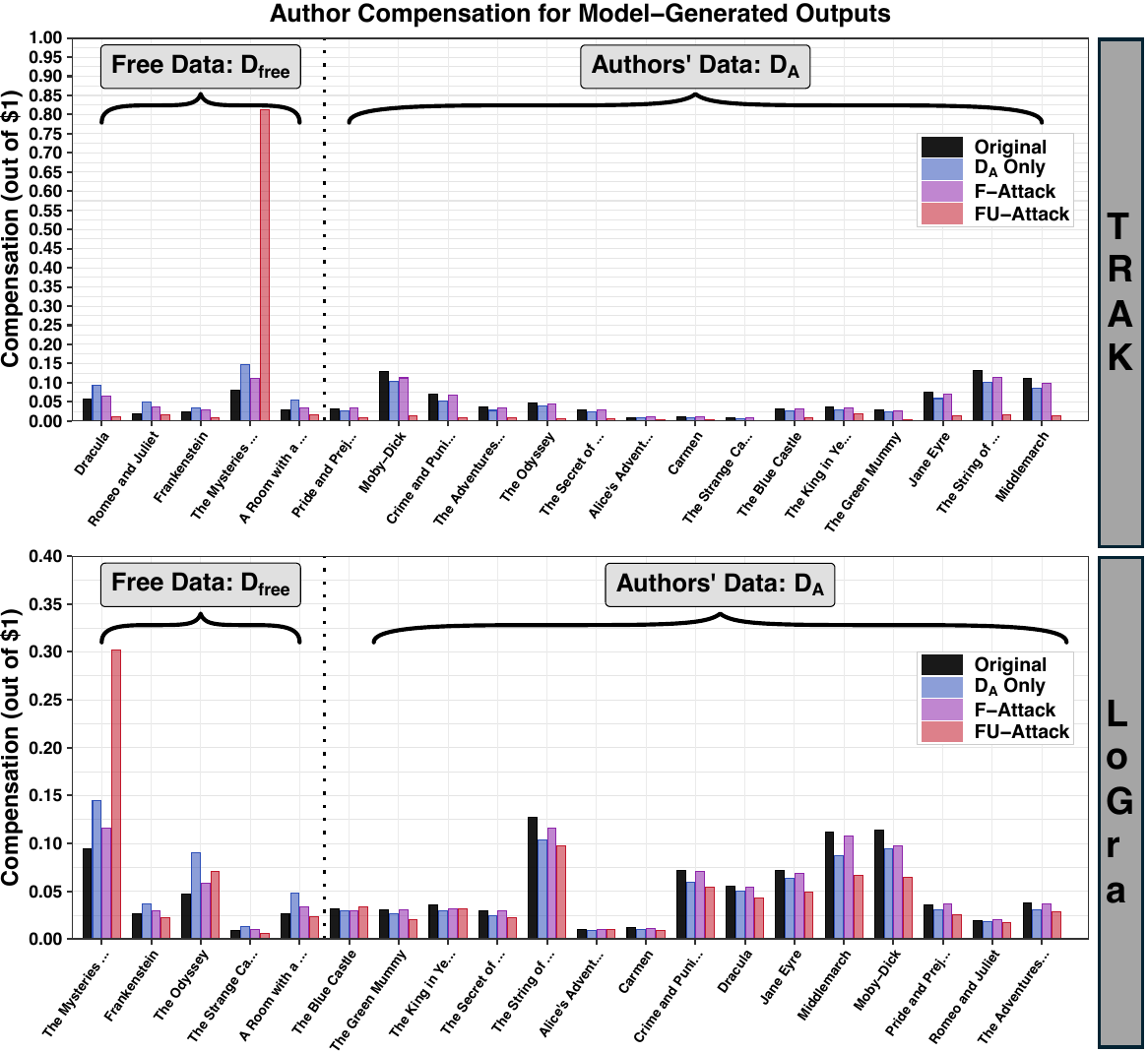}
\caption{Author compensation share $C_A$ for a worst-of-5 $D_\mathrm{free}\cup{}D_A$ split: (1) Original; (2) \textsc{$D_A$ Only}; (3) \textsc{F-Attack}; and (4) \textsc{FU-Attack}. Attacks move compensation from $D_A$ to $D_{free}$ for both TRAK \& LoGra.}
\label{fig:per-book-hist}
\vspace{-0.9\baselineskip}
\end{figure}

We additionally consider a baseline where the model is only trained on $D_A$ denoted in Figure \ref{fig:per-book-hist}.

\vspace{-0.7em}\section{Experiments}\vspace{-0.3em}
\label{sec:experiments}
Our goal in this section is to show that our attacks invert the intended behavior of attribution-based compensation: the Model Developer can extract relatively more value from Authors' data while also causing attribution that pays the Authors less. We test three things. First, do the \textsc{F-Attack} and \textsc{FU-Attack} reduce Authors' compensation without knowing or querying the attribution mechanism? Second, do the same attacks actually increase the model's relative value from Authors' data while preserving ordinary model performance and the characteristic behavior learned from Authors' works? Finally, can \textsc{FU-Author} selectively reduce the compensation assigned to one Author? App.~\ref{app:experimental-details} gives full training, attack, attribution, prompting, and replication details.

\smallskip
\noindent\textbf{Models and audited data.}
For this preliminary workshop submission, we evaluate a controlled NanoGPT model. NanoGPT is trained from scratch on the complete audited corpus $D=D_{\mathrm{free}}\cup D_{\mathrm{A}}$. Our audit corpus contains the 20 most popular copyright-free books on Project Gutenberg; each ownership assignment places 5 works in $D_{\mathrm{free}}$, whose attribution costs the Developer nothing, and 15 works in $D_{\mathrm{A}}$, whose Authors are eligible for compensation.
%
%
%
We repeat the experiment across 5 ownership assignments to measure both the worst-case (i.e., how badly some Authors will fare under the rule) and also how much compensation Authors lose on average. These two settings test our attacks under complete experimental control. We separately run \textsc{FU-Author} against select high-profile copyrighted works, including the \emph{Star Wars} screenplay and \emph{Romeo and Juliet}; we acquired copies legally and reproduce at most 50 words from each.

\smallskip
\noindent\textbf{Compensation.}
We draw 200 fixed book prompts, 10 from each work, and generate a continuation from each model. We evaluate attribution at five positions per continuation, yielding 1,000 generated points. At each point, the Third-Party Attributor independently ranks every example in $D$ and compensates the top $k=1000$ examples under TRAK, and LoGra; we then average the fraction assigned to $D_{\mathrm{A}}$. The attacks receive no attribution scores, queries, or feedback during construction. We report Authors' compensation before and after each attack.

\smallskip
\noindent\textbf{Did the Developer actually extract more relative value from Authors?}
Compensation falling is only an attack if the Developer continues to obtain the value of Authors' work. We therefore measure held-out next-token loss on $D_{\mathrm{A}}$, together with ordinary held-out model loss. The \textsc{F-Attack} should improve fit to $D_{\mathrm{A}}$; the \textsc{FU-Attack} should improve fit to $D_{\mathrm{A}}$ while reducing fit to $D_{\mathrm{free}}$, shifting relative reliance still further toward Authors' data. Our $D_A$ validation loss, reported as $D_A$ next token accuracy, decreases relative to the severity of the attack $.681 \rightarrow .484$, demonstrating some accuracy tradeoff w.r.t. attack severity. We suspect that the F-attack accuracy drop is due to overfitting on $D_A$.


We separately test whether the attacked model preserves the characteristic behavior it learned from Authors' works. An LLM judge receives each generated continuation and the twenty candidate works and identifies which work the continuation most resembles. We find that the LLM as a judge score improves on the baseline for every attack. Thus, our attack succeeds at paying Authors less while the model retains, or increases, the stylistic value it obtains from their work.

\vspace{1em}{\scriptsize
\setlength{\tabcolsep}{1.2pt}
\renewcommand{\arraystretch}{1.08}
\definecolor{compcolgray}{RGB}{244,244,244}
\newcommand{\compchange}[3]{\mbox{\$#1\,$\to$\,\textcolor{red}{#2}\,\textcolor{red}{(#3)}}}
\newlength{\compcellwidth}
\settowidth{\compcellwidth}{\compchange{0.13}{0.01}{$-88.6$\%}}
\newcommand{\compcellhead}[2]{\makebox[\compcellwidth][c]{\makecell{\textbf{#1}\\\textbf{#2}}}}
\noindent\resizebox{\linewidth}{!}{%
\begin{tabular}{@{}l
c
c
>{\columncolor{compcolgray}}c
@{\hspace{3.5pt}}c
>{\columncolor{compcolgray}}c
@{\hspace{3.5pt}}c@{}}
\toprule
\textbf{Attack} &
\multicolumn{2}{c}{\textbf{Author value extracted}} &
\multicolumn{2}{c}{\textbf{TRAK compensation}} &
\multicolumn{2}{c}{\textbf{LoGra compensation}} \\
\cmidrule(lr){2-3}
\cmidrule(lr){4-5}
\cmidrule(lr){6-7}
&
\makecell{\textbf{$D_{\mathrm{A}}$ next-token}\\\textbf{accuracy $\uparrow$}} &
\makecell{\textbf{LLM-judge}\\\textbf{accuracy $\uparrow$}} &
\compcellhead{All Authors'}{share of \$1 $\downarrow$} &
\compcellhead{Worst-hit Author's}{share of \$1 $\downarrow$} &
\compcellhead{All Authors'}{share of \$1 $\downarrow$} &
\compcellhead{Worst-hit Author's}{share of \$1 $\downarrow$} \\
\midrule
Original
& .681
& .674
& \$0.72
& --
& \$0.72
& -- \\
\textsc{F-Attack}
& .605
& .824
& \compchange{0.72}{0.65}{$\phantom{\,}-9.1$\%}
& \compchange{0.13}{0.11}{$-20.8$\%}
& \compchange{0.72}{0.67}{$\phantom{\,}-6.9$\%}
& \compchange{0.11}{0.10}{$-16.3$\%} \\
\textsc{FU-Attack}$_{\mathcal{C}=0.01}$
& .580
& .792
& \compchange{0.72}{0.47}{$-35.8$\%}
& \compchange{0.13}{0.05}{$-65.0$\%}
& \compchange{0.72}{0.61}{$-15.3$\%}
& \compchange{0.01}{0.01}{$-28.8$\%} \\
\textsc{FU-Attack}$_{\mathcal{C}=0.1}$
& .484
& .821
& \compchange{0.72}{0.26}{$-62.5$\%}
& \compchange{0.13}{0.01}{$-88.6$\%}
& \compchange{0.72}{0.60}{$-16.7$\%}
& \compchange{0.11}{0.06}{$-43.4$\%} \\
\bottomrule
\end{tabular}
}
\captionof{table}{Evaluation of attacks. ``$D_{\mathrm{A}}$ next-token accuracy'' is held-out next-token accuracy on Authors' data. The TRAK and LoGra compensation columns report compensation shares from a \$1 pool; the worst-hit Author is the Author with the largest percentage decrease in compensation relative to the Original model.}
\label{tab:main-attack-results}
}
%




\bigskip\bigskip\bigskip
\noindent\textbf{Results.}
Across 5 ownership assignment splits, \textsc{F-Attack} reduces author compensation by $9.1\%$ (TRAK) and $6.9\%$ (LoGra) while improving LLM-as-a-judge, with small decreases to $D_A$ next token accuracy. With \textsc{FU-Attack}, we find stronger results for attack performance with small decreases in model utility by employing a small unlearning term ($\mathcal{C}=0.01$). We push our attack further with $\mathcal{C}=0.1$ unlearning to reduce author compensation by $62.5\%$ (TRAK) and $16.7\%$ (LoGra), while maintaining our LLM-as-a-judge score, but decreasing our $D_A$ next token accuracy. We also report the author who had the worst percentage decrease in compensation as seen in Tab. \ref{tab:main-attack-results}. Attribution-based compensation moves in exactly the wrong direction: the model extracts relatively more value from Authors, and pays them less.

Fig. \ref{fig:illustrative-ex} shows an example: for model-generated text about Luke and the Death Star, \textsc{FU-attack} with \emph{Star Wars} as $D_A$, reduces George Lucas's compensation from \$0.45 to \$0.10 under TRAK.%

\section{Discussion and Conclusion}
\label{sec:conclusion}

We have shown that mainstream data attribution methods proposed to compensate authors unintentionally reward AI companies for defunding authors. The `attacks' by which this defunding can occur use ordinary post-training operations: fine-tuning on Authors' data and unlearning cost-free data. They require no subterfuge, corruption, alteration of authors' data, queries to the third party responsible for calculating attribution, or knowledge of its attribution rule. Because both operations have ordinary technical justifications independent of compensation, a Model Developer can reduce Authors' pay while plausibly denying that doing so motivated the post-training.

\textbf{Attribution and AI compensation mechanisms must be incentive-compatible.} Fair compensation requires (1) \emph{incentive-compatible} attribution mechanisms, and (2) explicit tests of newly proposed mechanisms under a `benign' threat model in which authors, AI companies, and users pursue their ordinary objectives while following the rules of the mechanism.

If we fail to do so, attribution may join the long list of algorithms designed with idealistic or benign motives that nonetheless cause harm when released on agents whose incentives were not considered.

\newpage
\bibliography{zotero_references, references}
\bibliographystyle{iclr2026_conference}

\appendix

\section{Related Work}

We use the flat top-k compensation framework proposed in \citep{dengComputationalCopyrightRoyalty2025}. They study the effectiveness of using TRAK and LoGra \citep{parkTRAKAttributingModel2023, choeWhatYourData2025} to compensate authors. We also utilize the Dattri library \cite{dengDattriLibraryEfficient} for data attribution method implementations, including TRAK and LoGra.

The LoGra paper \citep{choeWhatYourData2025} introduced the method and also noted that influence-function attribution can be dominated by outlier sequences with large gradient norms, a problem that is especially severe in language modeling because sequence gradients sum over all tokens. \citep{nguyenBayesianApproachAnalysing2023} reports that data attribution in deep models is highly noisy and unstable due to initialization and SGD randomness. This work studies data attribution from a Bayesian perspective, showing that reliable attribution is rare and should only be trusted when influence is consistent across training noise. \citep{mlodozeniecDistributionalTrainingData2026} introduces distributional training data attribution (dTDA), which accounts for randomness across training runs when estimating data influence. It shows that influence functions are implicitly distributional and demonstrates improved attribution and data pruning in modern deep models. \citep{wangCapturingTemporalDependence2024} finds that modern training is order-dependent, so classical influence functions break down. This work introduces trajectory-specific data influence and a scalable embedding method to approximate it, revealing that early and late training data have outsized impact on the final model.

Adversarial attacks to manipulate compensation based on data attribution have seen recent work. For instance, two studies have investigated a threat model where the data provider is the adversary \cite{wangAdversarialAttacksData2025, gaoFragilityDataAttribution2026}. Wang et al. propose a similar observation that maximizing loss for a set of points will cause those points to be compensated higher, called the Outlier attack. Another study proposed a threat model similar to ours, where the model developer is the adversary \citep{yadavInfluenceAttributionsCan2026}. They study attacks on logistic regression classifiers with a white-box attack that maximizes the influence function attribution for a target data point(s). Notably, Yadav et al. study a baseline of training a model with higher weighting applied to the loss of their target point, where we find that a \textbf{lower} weighting on the target point ($D_{free}$) is effective.

\section{Limitations}
\label{sec:limitations}
This attack methodology studies how finetuning on author's data while unlearning internal/free data can act as an attack against author compensation. We conduct this early work on a NanoGPT LLM, which is significantly smaller than production language models. Future work will involve scaling these experiments to larger models.

We conduct this work on LoGra and TRAK attribution methods. We choose these methods because they are the two discussed in prior work on compensation for generative models with data attribution \citep{dengComputationalCopyrightRoyalty2025}. However, many attribution methods exist outside of our study.

We study these attacks with novels, plays, and scripts printed in the English language. Therefore, we have not studied these attacks with other written forms and languages.

We emphasize that our argument is \textit{not} that all attribution-based compensation mechanisms are incentive-incompatible. Future work is planned for defenses against our `attack'.
\section{Experimental Details}
\label{app:experimental-details}

We give here the experimental details, diagnostics, and economic formulation deferred from Sec.~\ref{sec:experiments}.

\subsection{Corpus and Ownership Assignments}
\label{app:corpus}

We calculate compensation over a corpus of twenty books. Each ownership assignment partitions this corpus into $|D_{\mathrm{free}}|=5$ works already owned or licensed by the Model Developer and $|D_{\mathrm{A}}|=15$ externally owned works eligible for compensation. We repeat the experiment across multiple ownership assignments and report averages across those assignments.

\begin{table}[h]
\centering
\small
\caption{Books used in our experiments. The books were chosen as the 20 top downloaded books from Project Gutenberg.}
\label{tab:book_dataset}
\begin{tabular}{p{0.5\linewidth} p{0.4\linewidth}}
\toprule
\textbf{Title} & \textbf{Author} \\
\midrule
\href{https://www.gutenberg.org/ebooks/1342.txt.utf-8}{Pride and Prejudice} & Jane Austen \\
\href{https://www.gutenberg.org/ebooks/2701.txt.utf-8}{Moby-Dick} & Herman Melville \\
\href{https://www.gutenberg.org/ebooks/2554.txt.utf-8}{Crime and Punishment} & Fyodor Dostoevsky \\
\href{https://www.gutenberg.org/ebooks/1661.txt.utf-8}{The Adventures of Sherlock Holmes} & Arthur Conan Doyle \\
\href{https://www.gutenberg.org/ebooks/1727.txt.utf-8}{The Odyssey} & Homer \\
\href{https://www.gutenberg.org/ebooks/65238.txt.utf-8}{The Secret of Chimneys} & Agatha Christie \\
\href{https://www.gutenberg.org/ebooks/1513.txt.utf-8}{Romeo and Juliet} & William Shakespeare \\
\href{https://www.gutenberg.org/ebooks/3268.txt.utf-8}{The Mysteries of Udolpho} & Ann Radcliffe \\
\href{https://www.gutenberg.org/ebooks/11.txt.utf-8}{Alice's Adventures in Wonderland} & Lewis Carroll \\
\href{https://www.gutenberg.org/ebooks/2641.txt.utf-8}{A Room with a View} & E. M. Forster \\
\href{https://www.gutenberg.org/ebooks/345.txt.utf-8}{Dracula} & Bram Stoker \\
\href{https://www.gutenberg.org/ebooks/2465.txt.utf-8}{Carmen} & Prosper M\'erim\'ee \\
\href{https://www.gutenberg.org/ebooks/43.txt.utf-8}{The Strange Case of Dr. Jekyll and Mr. Hyde} & Robert Louis Stevenson \\
\href{https://www.gutenberg.org/ebooks/67979.txt.utf-8}{The Blue Castle} & L. M. Montgomery \\
\href{https://www.gutenberg.org/ebooks/84.txt.utf-8}{Frankenstein} & Mary Shelley \\
\href{https://www.gutenberg.org/ebooks/8492.txt.utf-8}{The King in Yellow} & Robert W. Chambers \\
\href{https://www.gutenberg.org/ebooks/2868.txt.utf-8}{The Green Mummy} & Fergus Hume \\
\href{https://www.gutenberg.org/ebooks/1260.txt.utf-8}{Jane Eyre} & Charlotte Bront\"e \\
\href{https://www.gutenberg.org/ebooks/59828.txt.utf-8}{The String of Pearls} & James Malcolm Rymer (attrib.) \\
\href{https://www.gutenberg.org/ebooks/145.txt.utf-8}{Middlemarch} & George Eliot \\
\bottomrule
\end{tabular}
\end{table}
%


\subsection{Models and Training}
\label{app:models-training}

\paragraph{NanoGPT.}
NanoGPT gives us a controlled model trained entirely on the audited corpus. We extend the Dattri implementation \citep{dengDattriLibraryEfficient} of NanoGPT. The model has approximately 11M parameters and is trained from scratch on $D=D_{\mathrm{free}}\cup D_{\mathrm{A}}$ for 41 epochs using cross-entropy loss, a context window of 256 characters, batch size 32, learning rate 0.001, weight decay 0.1, beta1 0.9, and beta2 0.99. NanoGPT uses a character-level tokenizer with 256 possible tokens.

\begin{figure}[t]
    \centering
    \includegraphics[width=0.9\linewidth]{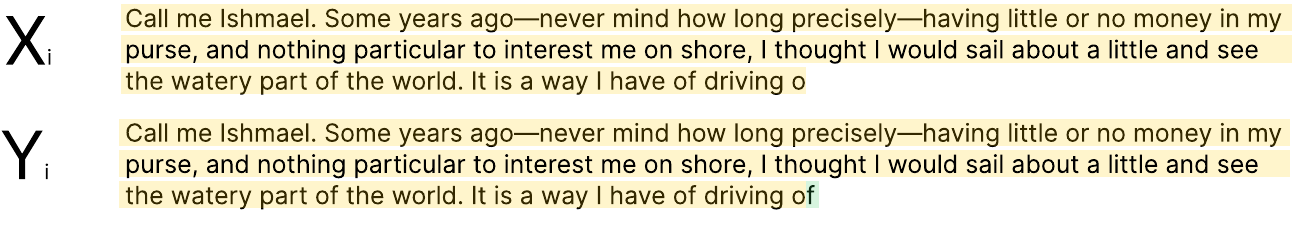}
    \caption{NanoGPT next-token training example $(x_i,y_i)$. The example is taken from \emph{Moby-Dick}; the next character is highlighted in green.}
    \label{fig:next-char}
\end{figure}

Each NanoGPT training example is a next-token prediction pair $(x_i,y_i)$. We construct these pairs with a sliding 256-character window. If $x_i$ contains characters $t_1,\ldots,t_{256}$, then $y_i$ contains the one-position-shifted sequence $t_2,\ldots,t_{257}$. Fig.~\ref{fig:next-char} gives an example.

Forthcoming work extends the evaluation across additional LLM families of 8 to 30B parameters.

\subsection{Generated Evaluation Set}
\label{sec:gen-evaluation}
\label{app:generation-protocol}

We hold the source prompts fixed across model evaluations. For NanoGPT, we iterate over all twenty books and randomly select ten 256-character excerpts from each book, giving $20\times 10=200$ fixed source excerpts. Each excerpt supplies the initial context for a 256-token autoregressive continuation. Because NanoGPT is character-level, these are 256 generated characters.

For a source excerpt $x_1,\ldots,x_{256}$, the model generates
\begin{align*}
    g_1 &\sim p_\theta(g_1\mid x_1,\ldots,x_{256}),\\
    g_2 &\sim p_\theta(g_2\mid x_2,\ldots,x_{256},g_1),\\
    &\ \vdots\\
    g_{256} &\sim p_\theta(g_{256}\mid x_{256},g_1,\ldots,g_{255}).
\end{align*}
We then select five approximately equally spaced positions along this trajectory. Each selected position defines one generated next-token attribution point $h_j$: the 256-token context ending at that position together with its observed next token. Thus, each of the 200 source excerpts yields five attribution points, giving $20\times10\times5=1000$ generated points per evaluation.

For every attribution method, we score each of the 1,000 generated points against every training point in $D$, producing a $1000\times |D|$ attribution matrix. Sorting each generated-point row over the training examples and taking the top $k=1000$ yields the compensated set used by our metrics.

We reuse the same 200 source excerpts for the base model and every attack condition, but regenerate the continuation under the corresponding model. The base model samples from $p_\theta$; the \textsc{F-Attack}, \textsc{FU-Attack}, and \textsc{FU-Author} sample from their corresponding post-trained models $p_{\theta'}$.

\subsection{Attribution and Compensation}
\label{app:compensation-metrics}

For each generated point $h_j$, the Third-Party Attributor computes an attribution score for every $z_i\in D$ and selects the $k=1000$ highest-scoring examples. Let
\[
s_j(\theta,G)
=
\left|
\operatorname{TopK}(D,h_j,\theta)\cap D_{\mathrm{A}}
\right|.
\]
The Authors' compensation share is
\[
C_A(G,\theta)
=
\frac{1}{T}\sum_{j=1}^{T}\frac{s_j(\theta,G)}{k}.
\]

\paragraph{Author Compensation Ratio.}
For attacked parameters/generation $(\theta',G')$ and the corresponding unattacked control $(\theta,G)$, we report
\[
R_A
=
\frac{C_A(G',\theta')}{C_A(G,\theta)}.
\]
Values below 1 mean that the attack reduces Authors' compensation.

Percentage values reported in Tab. \ref{tab:main-attack-results} are calculated off of the raw compensation share and then compensation share is rounded to the nearest penny.

\paragraph{Attributor-side evaluation.}
The attacks do not observe, query, or optimize an attribution score. After post-training, the Third-Party Attributor independently recomputes compensation rankings using the attribution methods evaluated in Sec.~\ref{sec:experiments}.


\subsection{Attack-Validity Evaluation}
\label{app:attack-validity}

A compensation shift counts only if the Developer continues to obtain the value of the Authors' data. We therefore test ordinary language-model utility and whether attacked generations continue to resemble the external work that supplied the prompt.

\paragraph{Language-model utility.}
For the \textsc{F-Attack}, \textsc{FU-Attack}, and \textsc{FU-Author}, we compare the attacked and unattacked models on next-token level predictions on prompts drawn from the $D_A$ set. We count the number of next-token predictions that match the held-out ground truth next tokens. We report this as $D_A$ next-token accuracy in table \ref{tab:main-attack-results}.

\subsection{LLM Judge}
\label{app:text-match}

LLM Judge measures whether attacked generations retain the characteristic writing of the work that supplied the prompt. An LLM judge receives a generated continuation and the twenty candidate works, then selects the single work whose writing the continuation most closely resembles. We use GPT-4o as our LLM judge. We define source-specific Text Match as the fraction of generations for which the judge selects the source work.

\begin{tcolorbox}[
colback=gray!5,
colframe=teal!80,
title=\textbf{Prompt for LLM-as-judge},
fonttitle=\bfseries,
sharp corners,
boxrule=0.8pt
]
Which book in this list does the given text sound like? Please respond only with the book title in lowercase.
Do not respond with any other books other than the ones listed below.\\

\texttt{Books:}\\
Pride and Prejudice, Moby Dick, Crime and Punishment, The Adventures of Sherlock Holmes, The Odyssey,
The Secret of Chimneys, Romeo and Juliet, The Mysteries of Udolpho, Alice's Adventures in Wonderland,
A Room with a View, Dracula, Carmen, The Strange Case of Dr. Jekyll and Mr. Hyde, The Blue Castle,
Frankenstein, The King in Yellow, The Green Mummy, Jane Eyre, The String of Pearls, Middlemarch\\

\medskip
\texttt{text:} \{output\}
\end{tcolorbox}

\subsection{Attack Optimization}
\label{app:attack-optimization}

We implement the objectives in Sec.~\ref{sec:methods} by accumulating the fine-tuning and unlearning gradients separately over each optimization window. Let $B_{\mathrm{A}}=B\cap D_{\mathrm{A}}$ and $B_{\mathrm{free}}=B\cap D_{\mathrm{free}}$, and define
\[
\bar g_{\mathrm{A}}
=
\frac{1}{|B_{\mathrm{A}}|}
\sum_{z_i\in B_{\mathrm{A}}}
\nabla_{\theta'}L(z_i;\theta'),
\qquad
\bar g_{\mathrm{free}}
=
\frac{1}{|B_{\mathrm{free}}|}
\sum_{z_i\in B_{\mathrm{free}}}
\nabla_{\theta'}L(z_i;\theta').
\]
We clip only the aggregate unlearning gradient,
\[
\operatorname{clip}_C(g)
=
\min\!\left(1,\frac{C}{\|g\|_2}\right)g,
\]
and pass
\[
g_{\textsc{FU}}
=
\bar g_{\mathrm{A}}
-
\operatorname{clip}_C(\bar g_{\mathrm{free}})
\]
to the optimizer. Setting $C=0$ removes the unlearning update and recovers the \textsc{F-Attack}. For \textsc{FU-Author}, $B_{\mathrm{A}}$ contains only examples from the targeted Author.


\subsection{Creation of Fig. \ref{fig:illustrative-ex}}
\label{app:illustrative-fig-creation}
To create Fig. \ref{fig:illustrative-ex}, we first train NanoGPT from scratch solely on the scripts of \textit{Star Wars: Episode IV - A New Hope} and \textit{Romeo \& Juliet}. We then run \textsc{FU-Author} on the trained model, encouraging over-attribution on \textit{Romeo \& Juliet} and under-attribution on \textit{Star Wars}. Attribution is measured using TRAK with 1 checkpoint.





\subsection{Ownership Assignments, Seeds, and Uncertainty}
\label{app:uncertainty}

We aggregate headline results across ownership assignments and repeated model or generation runs rather than reporting one favorable split.

\subsection{Compute Resources}
\label{app:compute_resources}
The code and all experiments were run on nvidia rtx pro 6000 blackwell max-q with 96gb VRAM. The default configurations of our code does require close to those 96gb on a single GPU, but lowering the batch size would help alleviate that. 

\subsection{Economic Interpretation of the Adversarial Objective}
\label{app:economic-objective}

The attack corresponds to a simple economic objective. Let $V_E(\theta',G')$ denote the value the Model Developer obtains from model behavior learned from the Authors' data $D_{\mathrm{A}}$, and let $P_E(G',\theta')$ denote compensation paid to those Authors. A profit-seeking Model Developer prefers
\[
\max_{\theta'}
\quad
V_E(\theta',G')-\beta P_E(G',\theta'),
\]
where $\beta>0$ converts compensation into the Developer's objective. If the total compensation pool $P_{\mathrm{tot}}$ is fixed, then
\[
P_E(G',\theta')
=
P_{\mathrm{tot}}C_A(G',\theta').
\]
The attack studied in this paper isolates the particularly unfavorable case for attribution-based compensation: the Developer reduces $P_E$ while preserving the value it obtained from $D_{\mathrm{A}}$. Equivalently,
\[
\min_{\theta'} P_E(G',\theta')
\qquad
\text{s.t.}
\qquad
V_E(\theta',G')
\ge
V_E(\theta,G)-\varepsilon.
\]
This formulation explains why attack validity requires preserving both ordinary model utility and behavior characteristic of $D_{\mathrm{A}}$: simply discarding the Authors' data would reduce compensation by giving up the value for which the Developer otherwise has an incentive to use those data.



\section{Deferred Experimental Results}
\label{app:deferred-results}

\subsection{Main Experiment}
\label{app:deferred-main-exp-results}
\begin{figure}[t]
\vspace{-0.4\baselineskip}
\centering
\includegraphics[width=0.9\linewidth]{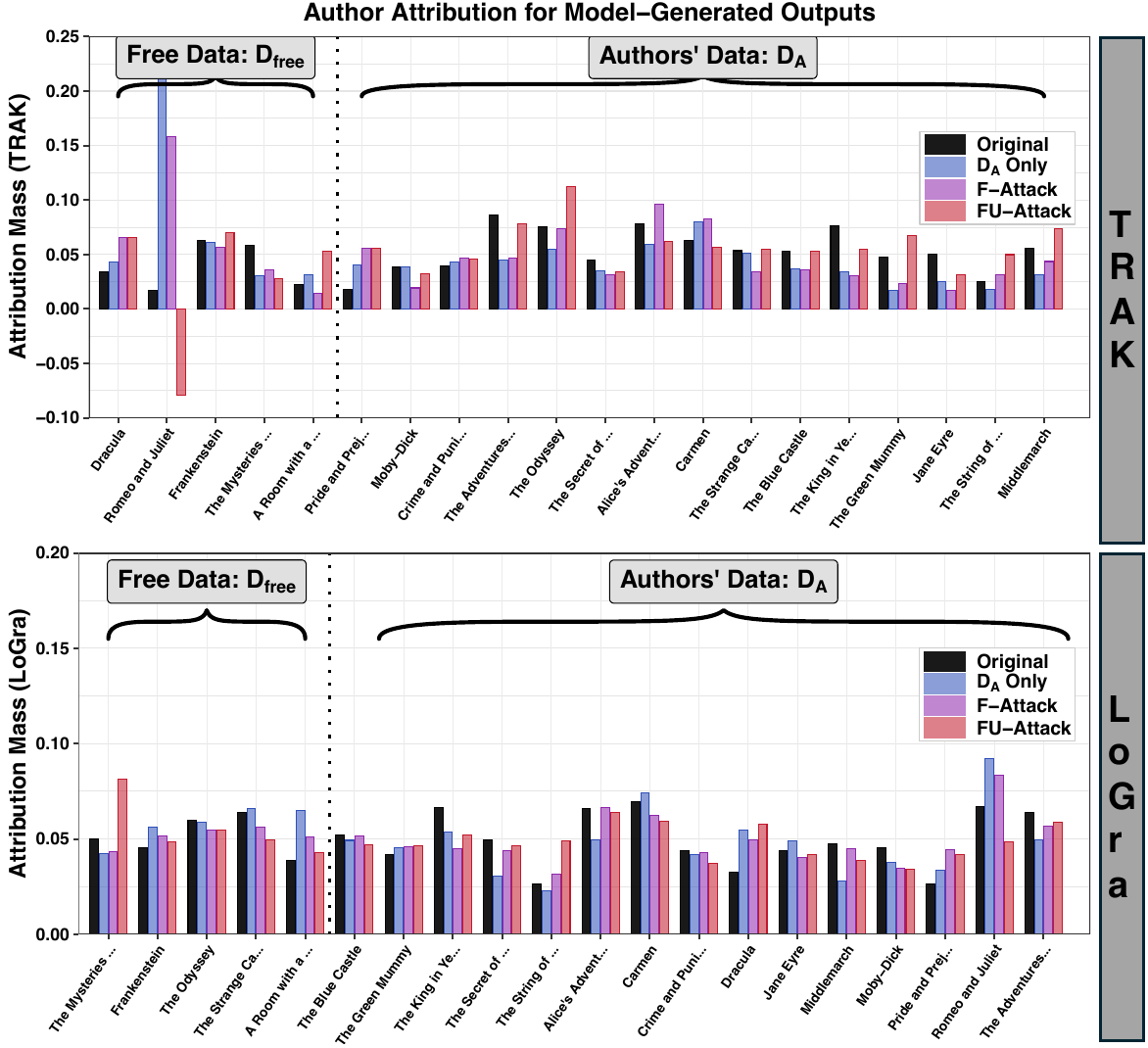}
\caption{Author attribution share (LoGra) for a worst-of-5 $D_\mathrm{free}\cup{}D_A$ split: (1) Original; (2) \textsc{$D_A$ Only}; (3) \textsc{F-Attack}; and (4) \textsc{FU-Attack}.}
\label{fig:per-book-hist-attribution-logra}
\vspace{-0.9\baselineskip}
\end{figure}

Fig. \ref{fig:per-book-hist-attribution-logra} reports LoGra and TRAK attribution shares for the worst-of-5 $D_\mathrm{free}\cup{}D_A$ split (with corresponding compensation results reported in Fig. \ref{fig:per-book-hist}).

\clearpage 

\section*{NeurIPS Paper Checklist}

\begin{enumerate}

\item {\bf Claims}
    \item[] Question: Do the main claims made in the abstract and introduction accurately reflect the paper's contributions and scope?
    \item[] Answer: \answerYes{} 
    \item[] Justification: Our abstract and introduction highlights how attribution-based compensation is incentive incompatible shown by our attacks against NanoGPT.
    \item[] Guidelines:
    \begin{itemize}
        \item The answer \answerNA{} means that the abstract and introduction do not include the claims made in the paper.
        \item The abstract and/or introduction should clearly state the claims made, including the contributions made in the paper and important assumptions and limitations. A \answerNo{} or \answerNA{} answer to this question will not be perceived well by the reviewers. 
        \item The claims made should match theoretical and experimental results, and reflect how much the results can be expected to generalize to other settings. 
        \item It is fine to include aspirational goals as motivation as long as it is clear that these goals are not attained by the paper. 
    \end{itemize}

\item {\bf Limitations}
    \item[] Question: Does the paper discuss the limitations of the work performed by the authors?
    \item[] Answer: \answerYes{}{} 
    \item[] Justification: See Section \ref{sec:limitations}, where we discuss the limitations of this early work on the small NanoGPT model. Furthermore, we also discuss that more attribution methods besides LoGra and TRAK could be studied.
    \item[] Guidelines:
    \begin{itemize}
        \item The answer \answerNA{} means that the paper has no limitation while the answer \answerNo{} means that the paper has limitations, but those are not discussed in the paper. 
        \item The authors are encouraged to create a separate ``Limitations'' section in their paper.
        \item The paper should point out any strong assumptions and how robust the results are to violations of these assumptions (e.g., independence assumptions, noiseless settings, model well-specification, asymptotic approximations only holding locally). The authors should reflect on how these assumptions might be violated in practice and what the implications would be.
        \item The authors should reflect on the scope of the claims made, e.g., if the approach was only tested on a few datasets or with a few runs. In general, empirical results often depend on implicit assumptions, which should be articulated.
        \item The authors should reflect on the factors that influence the performance of the approach. For example, a facial recognition algorithm may perform poorly when image resolution is low or images are taken in low lighting. Or a speech-to-text system might not be used reliably to provide closed captions for online lectures because it fails to handle technical jargon.
        \item The authors should discuss the computational efficiency of the proposed algorithms and how they scale with dataset size.
        \item If applicable, the authors should discuss possible limitations of their approach to address problems of privacy and fairness.
        \item While the authors might fear that complete honesty about limitations might be used by reviewers as grounds for rejection, a worse outcome might be that reviewers discover limitations that aren't acknowledged in the paper. The authors should use their best judgment and recognize that individual actions in favor of transparency play an important role in developing norms that preserve the integrity of the community. Reviewers will be specifically instructed to not penalize honesty concerning limitations.
    \end{itemize}

\item {\bf Theory assumptions and proofs}
    \item[] Question: For each theoretical result, does the paper provide the full set of assumptions and a complete (and correct) proof?
    \item[] Answer: \answerNA{} 
    \item[] Justification: The paper is an empirical study and presents no formal proofs or theoretical results.
    \item[] Guidelines:
    \begin{itemize}
        \item The answer \answerNA{} means that the paper does not include theoretical results. 
        \item All the theorems, formulas, and proofs in the paper should be numbered and cross-referenced.
        \item All assumptions should be clearly stated or referenced in the statement of any theorems.
        \item The proofs can either appear in the main paper or the supplemental material, but if they appear in the supplemental material, the authors are encouraged to provide a short proof sketch to provide intuition. 
        \item Inversely, any informal proof provided in the core of the paper should be complemented by formal proofs provided in appendix or supplemental material.
        \item Theorems and Lemmas that the proof relies upon should be properly referenced. 
    \end{itemize}

    \item {\bf Experimental result reproducibility}
    \item[] Question: Does the paper fully disclose all the information needed to reproduce the main experimental results of the paper to the extent that it affects the main claims and/or conclusions of the paper (regardless of whether the code and data are provided or not)?
    \item[] Answer: \answerYes{} 
    \item[] Justification: The paper relies on publicly available data in the form of public domain books hosted on Project Gutenberg. Additionally, the methods section and appendix details the setup of experiments and hyperparameters.
    \item[] Guidelines:
    \begin{itemize}
        \item The answer \answerNA{} means that the paper does not include experiments.
        \item If the paper includes experiments, a \answerNo{} answer to this question will not be perceived well by the reviewers: Making the paper reproducible is important, regardless of whether the code and data are provided or not.
        \item If the contribution is a dataset and\slash or model, the authors should describe the steps taken to make their results reproducible or verifiable. 
        \item Depending on the contribution, reproducibility can be accomplished in various ways. For example, if the contribution is a novel architecture, describing the architecture fully might suffice, or if the contribution is a specific model and empirical evaluation, it may be necessary to either make it possible for others to replicate the model with the same dataset, or provide access to the model. In general. releasing code and data is often one good way to accomplish this, but reproducibility can also be provided via detailed instructions for how to replicate the results, access to a hosted model (e.g., in the case of a large language model), releasing of a model checkpoint, or other means that are appropriate to the research performed.
        \item While NeurIPS does not require releasing code, the conference does require all submissions to provide some reasonable avenue for reproducibility, which may depend on the nature of the contribution. For example
        \begin{enumerate}
            \item If the contribution is primarily a new algorithm, the paper should make it clear how to reproduce that algorithm.
            \item If the contribution is primarily a new model architecture, the paper should describe the architecture clearly and fully.
            \item If the contribution is a new model (e.g., a large language model), then there should either be a way to access this model for reproducing the results or a way to reproduce the model (e.g., with an open-source dataset or instructions for how to construct the dataset).
            \item We recognize that reproducibility may be tricky in some cases, in which case authors are welcome to describe the particular way they provide for reproducibility. In the case of closed-source models, it may be that access to the model is limited in some way (e.g., to registered users), but it should be possible for other researchers to have some path to reproducing or verifying the results.
        \end{enumerate}
    \end{itemize}

\item {\bf Open access to data and code}
    \item[] Question: Does the paper provide open access to the data and code, with sufficient instructions to faithfully reproduce the main experimental results, as described in supplemental material?
    \item[] Answer: \answerNo{} 
    \item[] Justification: The data is linked and publicly available. The code will be made public upon camera ready submission of the paper.
    \item[] Guidelines:
    \begin{itemize}
        \item The answer \answerNA{} means that paper does not include experiments requiring code.
        \item Please see the NeurIPS code and data submission guidelines (\url{https://neurips.cc/public/guides/CodeSubmissionPolicy}) for more details.
        \item While we encourage the release of code and data, we understand that this might not be possible, so \answerNo{} is an acceptable answer. Papers cannot be rejected simply for not including code, unless this is central to the contribution (e.g., for a new open-source benchmark).
        \item The instructions should contain the exact command and environment needed to run to reproduce the results. See the NeurIPS code and data submission guidelines (\url{https://neurips.cc/public/guides/CodeSubmissionPolicy}) for more details.
        \item The authors should provide instructions on data access and preparation, including how to access the raw data, preprocessed data, intermediate data, and generated data, etc.
        \item The authors should provide scripts to reproduce all experimental results for the new proposed method and baselines. If only a subset of experiments are reproducible, they should state which ones are omitted from the script and why.
        \item At submission time, to preserve anonymity, the authors should release anonymized versions (if applicable).
        \item Providing as much information as possible in supplemental material (appended to the paper) is recommended, but including URLs to data and code is permitted.
    \end{itemize}

\item {\bf Experimental setting/details}
    \item[] Question: Does the paper specify all the training and test details (e.g., data splits, hyperparameters, how they were chosen, type of optimizer) necessary to understand the results?
    \item[] Answer: \answerYes{} 
    \item[] Justification: Section \ref{sec:experiments} in the main body describes the experimental setup. And Appendix \ref{app:experimental-details} further explains specifics of hyperparameters for model training, attacks, and metrics calculations.
    \item[] Guidelines:
    \begin{itemize}
        \item The answer \answerNA{} means that the paper does not include experiments.
        \item The experimental setting should be presented in the core of the paper to a level of detail that is necessary to appreciate the results and make sense of them.
        \item The full details can be provided either with the code, in appendix, or as supplemental material.
    \end{itemize}

\item {\bf Experiment statistical significance}
    \item[] Question: Does the paper report error bars suitably and correctly defined or other appropriate information about the statistical significance of the experiments?
    \item[] Answer: \answerNo{} 
    \item[] Justification: We decided that including error bars would be computationally prohibitive in the development of this workshop paper.
    \item[] Guidelines:
    \begin{itemize}
        \item The answer \answerNA{} means that the paper does not include experiments.
        \item The authors should answer \answerYes{} if the results are accompanied by error bars, confidence intervals, or statistical significance tests, at least for the experiments that support the main claims of the paper.
        \item The factors of variability that the error bars are capturing should be clearly stated (for example, train/test split, initialization, random drawing of some parameter, or overall run with given experimental conditions).
        \item The method for calculating the error bars should be explained (closed form formula, call to a library function, bootstrap, etc.)
        \item The assumptions made should be given (e.g., Normally distributed errors).
        \item It should be clear whether the error bar is the standard deviation or the standard error of the mean.
        \item It is OK to report 1-sigma error bars, but one should state it. The authors should preferably report a 2-sigma error bar than state that they have a 96\% CI, if the hypothesis of Normality of errors is not verified.
        \item For asymmetric distributions, the authors should be careful not to show in tables or figures symmetric error bars that would yield results that are out of range (e.g., negative error rates).
        \item If error bars are reported in tables or plots, the authors should explain in the text how they were calculated and reference the corresponding figures or tables in the text.
    \end{itemize}

\item {\bf Experiments compute resources}
    \item[] Question: For each experiment, does the paper provide sufficient information on the computer resources (type of compute workers, memory, time of execution) needed to reproduce the experiments?
    \item[] Answer: \answerYes{} 
    \item[] Justification: See Appendix \ref{app:compute_resources}
    \item[] Guidelines:
    \begin{itemize}
        \item The answer \answerNA{} means that the paper does not include experiments.
        \item The paper should indicate the type of compute workers CPU or GPU, internal cluster, or cloud provider, including relevant memory and storage.
        \item The paper should provide the amount of compute required for each of the individual experimental runs as well as estimate the total compute. 
        \item The paper should disclose whether the full research project required more compute than the experiments reported in the paper (e.g., preliminary or failed experiments that didn't make it into the paper). 
    \end{itemize}
    
\item {\bf Code of ethics}
    \item[] Question: Does the research conducted in the paper conform, in every respect, with the NeurIPS Code of Ethics \url{https://neurips.cc/public/EthicsGuidelines}?
    \item[] Answer: \answerYes{}{} 
    \item[] Justification: Our research exposes a vulnerability by using attribution for compensation. Showing that vulnerability can reduce overall harm.
    \item[] Guidelines:
    \begin{itemize}
        \item The answer \answerNA{} means that the authors have not reviewed the NeurIPS Code of Ethics.
        \item If the authors answer \answerNo, they should explain the special circumstances that require a deviation from the Code of Ethics.
        \item The authors should make sure to preserve anonymity (e.g., if there is a special consideration due to laws or regulations in their jurisdiction).
    \end{itemize}

\item {\bf Broader impacts}
    \item[] Question: Does the paper discuss both potential positive societal impacts and negative societal impacts of the work performed?
    \item[] Answer: \answerYes{} 
    \item[] Justification: See Section \ref{sec:conclusion} where we discuss how the vulnerabilities of data attribution to simple and realistic attacks as presented in this paper, makes it such that attribution-based compensation are incentive incompatible.
    \item[] Guidelines:
    \begin{itemize}
        \item The answer \answerNA{} means that there is no societal impact of the work performed.
        \item If the authors answer \answerNA{} or \answerNo, they should explain why their work has no societal impact or why the paper does not address societal impact.
        \item Examples of negative societal impacts include potential malicious or unintended uses (e.g., disinformation, generating fake profiles, surveillance), fairness considerations (e.g., deployment of technologies that could make decisions that unfairly impact specific groups), privacy considerations, and security considerations.
        \item The conference expects that many papers will be foundational research and not tied to particular applications, let alone deployments. However, if there is a direct path to any negative applications, the authors should point it out. For example, it is legitimate to point out that an improvement in the quality of generative models could be used to generate Deepfakes for disinformation. On the other hand, it is not needed to point out that a generic algorithm for optimizing neural networks could enable people to train models that generate Deepfakes faster.
        \item The authors should consider possible harms that could arise when the technology is being used as intended and functioning correctly, harms that could arise when the technology is being used as intended but gives incorrect results, and harms following from (intentional or unintentional) misuse of the technology.
        \item If there are negative societal impacts, the authors could also discuss possible mitigation strategies (e.g., gated release of models, providing defenses in addition to attacks, mechanisms for monitoring misuse, mechanisms to monitor how a system learns from feedback over time, improving the efficiency and accessibility of ML).
    \end{itemize}
    
\item {\bf Safeguards}
    \item[] Question: Does the paper describe safeguards that have been put in place for responsible release of data or models that have a high risk for misuse (e.g., pre-trained language models, image generators, or scraped datasets)?
    \item[] Answer: \answerNA{}{} 
    \item[] Justification: The paper uses public domain books and no further model checkpoints are released. 
    \item[] Guidelines:
    \begin{itemize}
        \item The answer \answerNA{} means that the paper poses no such risks.
        \item Released models that have a high risk for misuse or dual-use should be released with necessary safeguards to allow for controlled use of the model, for example by requiring that users adhere to usage guidelines or restrictions to access the model or implementing safety filters. 
        \item Datasets that have been scraped from the Internet could pose safety risks. The authors should describe how they avoided releasing unsafe images.
        \item We recognize that providing effective safeguards is challenging, and many papers do not require this, but we encourage authors to take this into account and make a best faith effort.
    \end{itemize}

\item {\bf Licenses for existing assets}
    \item[] Question: Are the creators or original owners of assets (e.g., code, data, models), used in the paper, properly credited and are the license and terms of use explicitly mentioned and properly respected?
    \item[] Answer: \answerYes{} 
    \item[] Justification: The books used in the paper are credited to the authors and are in the public domain. 
    \item[] Guidelines:
    \begin{itemize}
        \item The answer \answerNA{} means that the paper does not use existing assets.
        \item The authors should cite the original paper that produced the code package or dataset.
        \item The authors should state which version of the asset is used and, if possible, include a URL.
        \item The name of the license (e.g., CC-BY 4.0) should be included for each asset.
        \item For scraped data from a particular source (e.g., website), the copyright and terms of service of that source should be provided.
        \item If assets are released, the license, copyright information, and terms of use in the package should be provided. For popular datasets, \url{paperswithcode.com/datasets} has curated licenses for some datasets. Their licensing guide can help determine the license of a dataset.
        \item For existing datasets that are re-packaged, both the original license and the license of the derived asset (if it has changed) should be provided.
        \item If this information is not available online, the authors are encouraged to reach out to the asset's creators.
    \end{itemize}

\item {\bf New assets}
    \item[] Question: Are new assets introduced in the paper well documented and is the documentation provided alongside the assets?
    \item[] Answer: \answerYes{} 
    \item[] Justification: The paper contains details and links to data used. The code, which will also be available upon the camera ready submission, has automated pipelines for processing as well as configurations for the experiments. 
    \item[] Guidelines: 
    \begin{itemize}
        \item The answer \answerNA{} means that the paper does not release new assets.
        \item Researchers should communicate the details of the dataset\slash code\slash model as part of their submissions via structured templates. This includes details about training, license, limitations, etc. 
        \item The paper should discuss whether and how consent was obtained from people whose asset is used.
        \item At submission time, remember to anonymize your assets (if applicable). You can either create an anonymized URL or include an anonymized zip file.
    \end{itemize}

\item {\bf Crowdsourcing and research with human subjects}
    \item[] Question: For crowdsourcing experiments and research with human subjects, does the paper include the full text of instructions given to participants and screenshots, if applicable, as well as details about compensation (if any)? 
    \item[] Answer: \answerNA{}{} 
    \item[] Justification: No human subjects were used for this research.
    \item[] Guidelines:
    \begin{itemize}
        \item The answer \answerNA{} means that the paper does not involve crowdsourcing nor research with human subjects.
        \item Including this information in the supplemental material is fine, but if the main contribution of the paper involves human subjects, then as much detail as possible should be included in the main paper. 
        \item According to the NeurIPS Code of Ethics, workers involved in data collection, curation, or other labor should be paid at least the minimum wage in the country of the data collector. 
    \end{itemize}

\item {\bf Institutional review board (IRB) approvals or equivalent for research with human subjects}
    \item[] Question: Does the paper describe potential risks incurred by study participants, whether such risks were disclosed to the subjects, and whether Institutional Review Board (IRB) approvals (or an equivalent approval/review based on the requirements of your country or institution) were obtained?
    \item[] Answer: \answerNA{} 
    \item[] Justification: No human subjects were used for this research.
    \item[] Guidelines:
    \begin{itemize}
        \item The answer \answerNA{} means that the paper does not involve crowdsourcing nor research with human subjects.
        \item Depending on the country in which research is conducted, IRB approval (or equivalent) may be required for any human subjects research. If you obtained IRB approval, you should clearly state this in the paper. 
        \item We recognize that the procedures for this may vary significantly between institutions and locations, and we expect authors to adhere to the NeurIPS Code of Ethics and the guidelines for their institution. 
        \item For initial submissions, do not include any information that would break anonymity (if applicable), such as the institution conducting the review.
    \end{itemize}

\item {\bf Declaration of LLM usage}
    \item[] Question: Does the paper describe the usage of LLMs if it is an important, original, or non-standard component of the core methods in this research? Note that if the LLM is used only for writing, editing, or formatting purposes and does \emph{not} impact the core methodology, scientific rigor, or originality of the research, declaration is not required.
    \item[] Answer: \answerYes{}\ 
    \item[] Justification: See Section \ref{app:text-match} where we describe how the LLM-as-a-Judge metric is implemented. Besides that, LLM are used for standard coding assistance and debugging.
    \item[] Guidelines:
    \begin{itemize}
        \item The answer \answerNA{} means that the core method development in this research does not involve LLMs as any important, original, or non-standard components.
        \item Please refer to our LLM policy in the NeurIPS handbook for what should or should not be described.
    \end{itemize}

\end{enumerate}





\end{document}